\documentclass[sigconf]{acmart}

\usepackage{acronym}

\begin{document}
\settopmatter{printacmref=false} 
\renewcommand\footnotetextcopyrightpermission[1]{} 
\pagestyle{plain} 
\newacro{GA}{Genetic Algorithm}
\newacro{GI}{Genetic Improvement}
\newacro{GP}{Genetic Programming}
\newacro{LGP}{Linear Genetic Programming}
\newacro{SSP}{Subset Sum Problem}
\newacro{-O3}{The default LLVM optimization level `-O3'}
\newacro{Pass}{LLVM Compiler Optimization Pass}

\title{Genetic Improvement in the Shackleton Framework for Optimizing LLVM Pass Sequences}

\author{Shuyue Stella Li}
\email{sli136@jhu.edu}
\affiliation{%
  \institution{Department of Computer Science, Johns Hopkins University}
  \city{Baltimore, MD}
  \country{USA}
}

\author{Hannah Peeler}
\authornote{At time of publication, no longer affiliated with Arm. Reachable at hpeeler@utexas.edu}
\email{hpeeler@utexas.edu}
\affiliation{%
  \institution{Arm Ltd.}
  \country{USA}
}

\author{Andrew N. Sloss}
\authornote{Now at University of Washington, Seattle, WA, USA.}
\email{andrew@sloss.net}
\affiliation{%
  \institution{Arm Ltd.}
  \country{USA}
}

\author{Kenneth N. Reid}
\email{ken@kenreid.co.uk}
\affiliation{%
 \institution{Department of Animal Science, Michigan State University}
 \city{East Lansing, MI}
 \country{USA}
}


\author{Wolfgang Banzhaf}
\email{banzhafw@msu.edu}
\affiliation{%
  \institution{Department of CSE, Michigan State University}
  \city{East Lansing, MI}
  \country{USA}
}

\renewcommand{\shortauthors}{Li and Peeler, et al.}

\begin{abstract}
  Genetic improvement is a search technique that aims to improve a given acceptable solution to a problem. In this paper, we present the novel use of genetic improvement to find problem-specific optimized LLVM pass sequences. We develop a pass-level patch representation in the linear genetic programming framework, Shackleton, to evolve the modifications to be applied to the default optimization pass sequences. Our GI-evolved solution has a mean of 3.7\% runtime improvement compared to the -O3 optimization level in the default code generation options which optimizes on runtime. The proposed GI method provides an automatic way to find a problem-specific optimization sequence that improves upon a general solution without any expert domain knowledge. In this paper, we discuss the advantages and limitations of the GI feature in the Shackleton Framework and present our results.
\end{abstract}



\maketitle

\section{Introduction} \label{section:introduction}
    \ac{GI} \cite{langdon2014optimizing, haraldsson2021genetic} automatically improves upon a given solution using \ac{GP} \cite{koza1992,bnkf, langdon2002foundations}. This approach is inspired by the process of natural selection 
    \cite{darwin1859origin}, in which the idea of relative fitness advantage 
    allows for the preservation of favorable variations and guides the passing of genetic information to the next generation. The \ac{GA} \cite{holland1975,goldberg1989genetic} is a powerful search algorithm that can efficiently find the near-optimal solution in a large search space. \ac{LGP} 
    \cite{bb2007} is a special application of its variant, Genetic Programming, in which the genetic information of each individual codes for active elements in the population represented in a sequential order. The Shackleton Framework\footnote{\url{https://github.com/ARM-software/Shackleton-Framework}} is a generalized \ac{LGP} framework that allows the use of \ac{GP} on any user-defined object types and fitness metrics \cite{peeler2022optimizing}. 

    In the \ac{GI} feature of the Shackleton Framework (Shackleton-GI), each modification from the baseline is represented as a sequence of operations to be applied to the baseline solution. The use-case of interest in our experiments is the optimization of LLVM\footnote{\url{https://llvm.org}} compiler optimization pass sequences \cite{lattner2006introduction}. LLVM is a collection of modular and reusable (language/target independent) compiler and tool-chain technologies. Different compile-time optimizations can be specified using LLVM (Transform) Passes, which traverse the program in some way and mutate the program in order to optimize some metric (i.e. reduce runtime)\cite{sarda2015llvm}; a sequence of LLVM passes can be specified at compilation to achieve a particular optimization goal. Shackleton-GI evolves a series of insertion, deletion, and replacement patch operations, which produces a more powerful optimization pass sequence when applied to a pre-defined sequence of LLVM passes, which is usually a solution to a general problem.


\section{Methods} \label{section:methods}

    Shackleton \cite{peeler2022optimizing} is a flexible \ac{LGP} framework, in which various types of objects can be treated as genes and optimized with the \ac{GA}. In Shackleton-GI, we develop a pass-level patch representation in which individuals consist of `genes' that are patches. A patch has a type field, a position field, and a value field. 
    The framework takes in a specific source code of a program, and generates a sequence of patches that will be used to modify a starting sequence of compiler optimization passes. 
    
\begin{figure}[H]
        \includegraphics[width=0.9\linewidth]{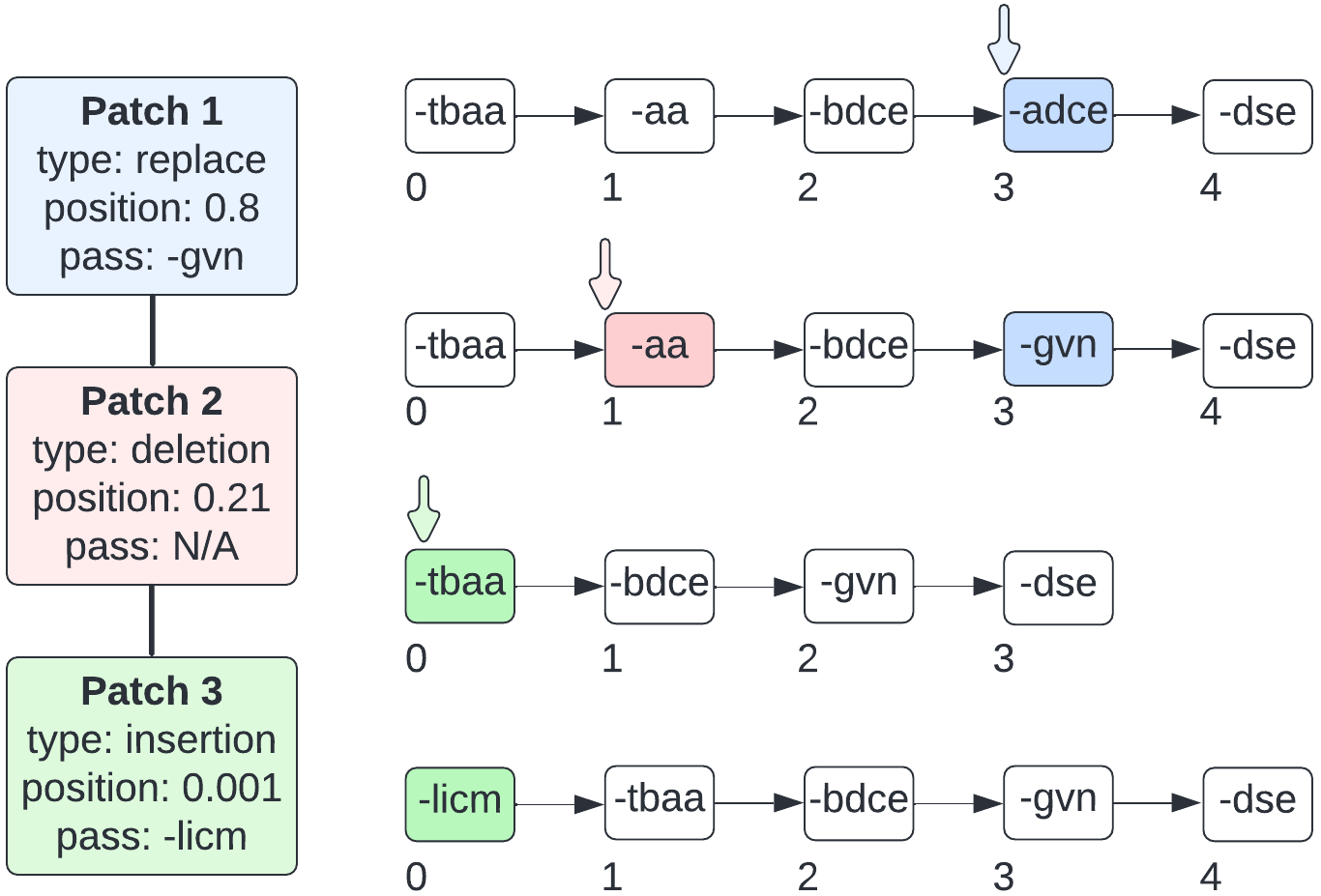}
        \caption{Sample Patch Representation of \ac{GI}. Three different patches are applied: insertion (1), deletion (2), and replacement (3). Position fields are relative positions between 0 and 1; value fields are LLVM pass names.}
        \label{fig:patch}
    \end{figure}

    A demonstration of the process is shown in Figure \ref{fig:patch}, in which an individual with three patches is applied to an example initial sequence of 5 LLVM passes long. After the initial sequence is modified by the patches contained in an individual, it is used during compilation as the optimization arguments. After the source program is compiled with the new pass sequence, the average runtime over 40 runs is recorded as the fitness value of that individual. This minimizes the effect of runtime inconsistency due to system fluctuations and any unusual halting in the user-provided source program.
    
    In our experiments, the source program is the Backtrack Algorithm implementation for the \ac{SSP}\footnote{\url{https://github.com/parthnan/SubsetSum-BacktrackAlgorithm}}, and the to-be-modified sequence is the LLVM optimization passes in the default LLVM optimization level -O3, which enables optimizations that take longer to perform during compilation or that may generate larger code in an attempt to make the program run faster \cite{lattner2006introduction, lattner2008llvm}. There are a number of hyperparameters required for Shackleton, and we used the optimal hyperparameter combinations found in \cite{peeler2022optimizing}.
    The experiments were conducted on HPCC nodes running CentOS Linux version 7 and Clang version 8.0.0.


\section{Results and Discussion} \label{section:results}

\begin{figure}[H]
        \includegraphics[width=\linewidth]{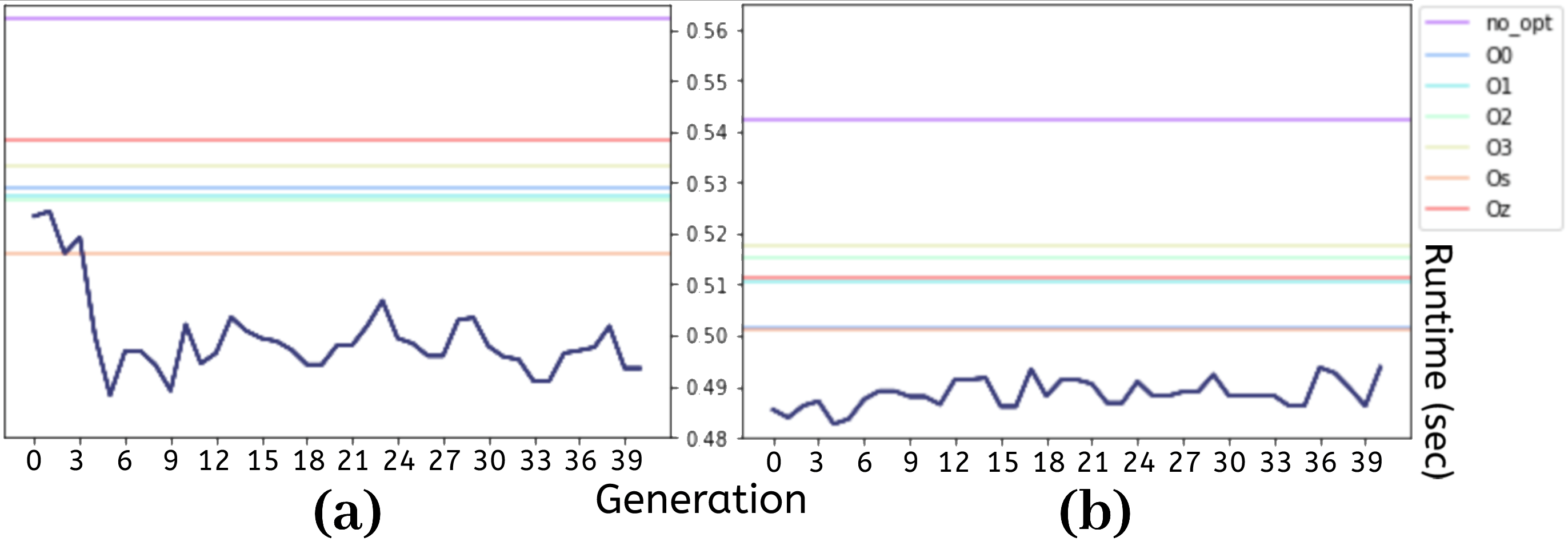}
    \caption{Sample Runtime Improvement}
    \label{fig:plots}
    \end{figure}

    Eight repeated trials were run with the same hyperparameter values 
    and the fitness across generations for two sample runs are plotted in Figure \ref{fig:plots}, with horizontal lines as the baseline runtime. As we can see, the run on the left shows a converging pattern that starts at a high runtime then decreases; the run on the right initializes at a high quality starting point, and stays within the same range during the entire evolutionary process. However, both scenarios are improvements upon the baselines.
    
    The average percent improvement compared to the LLVM default optimization level -O3 in terms of target program runtime over 8 runs of Shackleton-GI is 3.7\% with a standard deviation of 0.8768. The p-value for the left-tail test when the null hypothesis for the mean percent improvement of 0 is 0.000012\%. This shows the robustness of the algorithm and its readiness to be experimented with in production. 
    
    The search space for LLVM optimization pass sequences is in the order of $10^{167}$ (using 120 different passes in sequences of approximately 80 passes long). Therefore, finding the absolute optimum for a given source code is computationally impossible for currently available methods. The default -O3 optimization level in LLVM is carefully hand-crafted and aims at reducing the runtime of a target program. Hence, it gives a good starting point for the search and significantly reduces the size of the search space. The pass-level patch representation in Shackleton-GI effectively searches near this initial starting point and is able to find a local minimum tailored to the specific source program. The use of \ac{GI} significantly increases the efficiency of the search compared to a run from random solutions \cite{peeler2022optimizing} and is able to provide a better solution than -O3 as-is. 

\section{Conclusion and Future Work} \label{section:conclusion}

    The -O3 LLVM optimization level is hand-crafted by experts with rich domain-specific knowledge about the LLVM infrastructure, and is general enough to be used by different programs. Shackleton-GI automatically produces a sequence of patches that generates a problem-specific optimization solution to a user-provided source program. We proposed a pass-level patch representation for \ac{GI} that can be extended into different object types, and showed that our approach is able to achieve substantial runtime improvements compared to a strong compiler baseline.

    Shackleton-GI is a novel application of \ac{GI} and a first step in exploring a flexible use case of Shackleton. Future directions in the development of Shackleton-GI are: First, measuring fitness of individuals with clock speed, which could potentially be influenced by resource sharing on the same computing cluster. A more accurate measure would be to measure the CPU time by altering the threading design in the Shackleton Framework. Second, our experiments used the optimal hyperparameter values found by \cite{peeler2022optimizing} in a \ac{LGP} (non-GI) environment. Additional hyperparameter tuning might result in further runtime improvements as this is a different use case. Further investigation into other \ac{GI} algorithms and a wider range of test problems would also be interesting areas of future research.

\section{Acknowledgements}
    This work was generously funded by the EnSURE (Engineering Summer Undergraduate Research Experience) program at Michigan State University as well as the John R. Koza Endowment. We also gratefully acknowledge The Institute of Cyber-Enabled Research (ICER) at MSU for providing the hardware infrastructure that made the computation required to make this work possible. 

\bibliographystyle{ACM-Reference-Format}
\bibliography{references}
\end{document}